\documentclass{article}

\usepackage{PRIMEarxiv}

\usepackage[utf8]{inputenc}
\usepackage[T1]{fontenc}
\usepackage{hyperref}
\usepackage{url}
\usepackage{booktabs}
\usepackage{amsmath,amssymb,amsfonts}
\usepackage{microtype}
\usepackage{graphicx}
\usepackage{caption}
\usepackage{multirow}
\usepackage{xcolor}
\usepackage{pifont}

\graphicspath{{Figures/}{Figures/raw/}}
\title{Forwardrobe: Garment-Aware Gaussian Avatars from a Single Image}

\author{
  Daisheng Jin  \\
  Nanyang Technological University \\
  Singapore
  \\
    \And
  Shuyun Wang \\
  The University of Queensland \\
  Australia \\
   \And
  Ying He\thanks{Corresponding author: Y. He (Email: yhe@ntu.edu.sg)} \\
  Nanyang Technological University \\
  Singapore \\
}

\begin{document}
\maketitle

\begin{figure}[!htbp]
    \centering
    \includegraphics[width=0.90\linewidth]{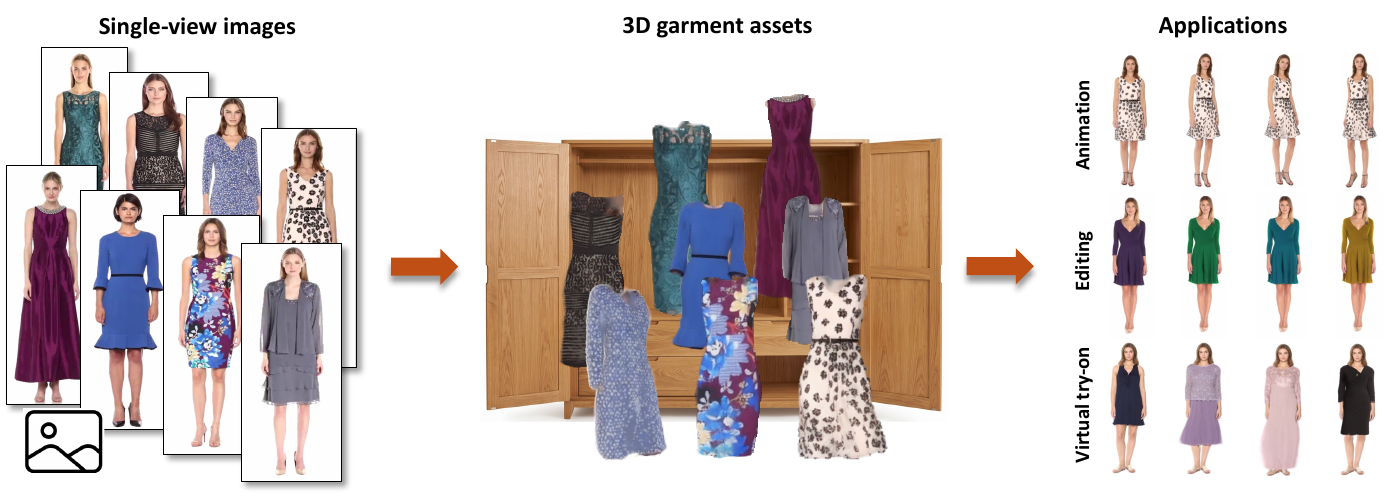}
    \caption{\textbf{Forwardrobe} reconstructs garment-aware Gaussian avatars from single-view images, improving the continuity and animation of loose garments. The explicitly separated garment layer forms an independently controllable 3D asset, supporting animation, appearance editing, and 3D virtual try-on across different avatars.}
    \label{fig:teaser}
\end{figure}

\begin{abstract}
Reconstructing animatable 3D human avatars from a single image remains particularly challenging for loose garments, whose geometry and motion cannot be adequately represented by body-aligned topology and skinning. We present \textbf{Forwardrobe}, a feed-forward framework for reconstructing garment-aware Gaussian avatars from a single image.  Forwardrobe explicitly separates clothing from the body in canonical Gaussian space and equips the garment layer with continuity-aware geometry and skinning initialization, pose-conditioned non-rigid deformation, and appearance adaptation. These designs improve garment reconstruction and visual quality during animation, particularly for skirts and dresses. The separated garment layer additionally forms an independently controllable 3D asset, enabling garment editing, transfer, and 3D virtual try-on. Experiments demonstrate improved garment reconstruction quality and greater flexibility in garment manipulation compared with existing single-image avatar reconstruction methods.

\end{abstract}

\keywords{3D human avatars \and 3D Gaussian Splatting \and garment reconstruction \and virtual try-on}

\section{Introduction}

High-quality 3D human avatars have become an important foundation for virtual presenters, film production, and immersive AR/VR applications. Traditionally, creating personalized avatars relies on professional artists or specialized multi-camera capture systems. While these pipelines can achieve impressive realism, their high acquisition cost and labor-intensive production process greatly limit accessibility and large-scale customization.

Recent advances in neural radiance fields~\cite{mildenhall2020nerf} and 3D Gaussian Splatting~\cite{kerbl20233d} have enabled high-fidelity reconstruction of human heads~\cite{xu2024gaussianhead,jin2025sfdm} and full-body avatars~\cite{peng2021animatablenerf,wang2022arah} from multi-view captures.
NeRF-based methods have demonstrated high-fidelity reconstruction and modeled pose-driven deformation in canonical space~\cite{zheng2023avatarrex}, while 3D Gaussian-based approaches~\cite{li2024animatablegaussians} enable more efficient optimization and real-time rendering.
Layered methods such as LayGA~\cite{lin2024layga} and Gaussian Wardrobe~\cite{chen2026gaussianwardrobe} further separate the body and clothing to support garment transfer and virtual try-on. However, these approaches typically require calibrated multi-view videos and costly capture setups.
To reduce these acquisition requirements, monocular-video approaches~\cite{guo2023vid2avatar} exploit motion and viewpoint variation within a single video to optimize personalized avatars. Although they achieve high visual quality and improved garment modeling, they still depend on time-consuming subject-specific optimization and sufficient coverage of poses, viewpoints, and body regions.

More recently, large-scale generative priors have made single-image human reconstruction increasingly practical. Methods such as PIFuHD~\cite{saito2020pifuhd} and ECON~\cite{xiu2023econ} reconstruct clothed geometry using parametric body models, normal estimation, or other human priors, while LHM~\cite{qiu2025lhm} and DynaAvatar~\cite{kwon2026dynaavatar} extend this setting toward animatable 3D Gaussian avatars and dynamic clothing. 
Despite substantial improvements in efficiency and generalization, most existing methods still represent the clothed human as a monolithic avatar, where the body, garments, appearance, and motion are entangled within a unified representation. Consequently, they provide limited garment-level controllability for applications such as the animation of loose garments, clothing transfer, and garment editing.

To address these limitations, we propose \textbf{Forwardrobe}, a feed-forward framework for reconstructing garment-aware Gaussian avatars from a single image. Forwardrobe leverages human-centric foundation models to initialize 3D geometry, garment masks, and a static avatar, then extracts garment Gaussians through a Garment Labeling Module. Language-level garment cues guide category-aware garment continuity and skinning initialization, using smoothly blended lower-body skinning for skirt-like garments while preserving separate leg articulation for pants-like garments. A Garment Dynamic Module further combines LBS-based coarse motion with pose-dependent geometry and appearance residuals, enabling independent modeling of garment structure, deformation, and shading. Complementary geometry, segmentation, contact, and appearance priors preserve coherent boundaries, stable textures, and natural body--garment integration during animation and transfer.

Unlike existing single-image methods that reconstruct a monolithic clothed avatar, Forwardrobe explicitly recovers an editable and transferable garment representation. It performs garment decoupling without requiring monocular videos or multi-view captures, and directly models garment geometry, motion, and appearance using unstructured 3D Gaussians. Forwardrobe therefore unifies single-image avatar reconstruction, garment manipulation, animation, and 3D virtual try-on within an efficient Gaussian avatar framework.

Our contributions are summarized as follows:

\begin{itemize}

\item To our knowledge, Forwardrobe is the first feed-forward framework to reconstruct the observed garment from a single clothed-person image as an independent, motion-conditioned 3D Gaussian asset and recompose it with another avatar without per-garment optimization.

\item We introduce a garment-aware representation that combines continuity-aware geometry and skinning initialization with pose-conditioned deformation and appearance adaptation, improving garment quality and visual continuity, particularly for skirt-like garments.

\item Experiments demonstrate improved garment reconstruction and animation quality. The separated representation further enables garment editing and cross-avatar 3D virtual try-on within a shared canonical template.

\end{itemize}

\section{Related Work}

\subsection{3D Animatable Human Avatars}

To animate digital humans under novel poses, animatable avatar methods typically construct a human representation in canonical space and deform it into the posed observation space using LBS, inverse skinning, or pose-conditioned deformation. Implicit-field methods, such as Animatable NeRF~\cite{peng2021animatablenerf} and ARAH~\cite{wang2022arah}, combine canonical radiance fields or signed distance fields (SDF) with pose-driven deformation fields to support novel-view and novel-pose synthesis. However, they often require costly training and rendering, and may struggle with large pose changes, loose clothing, and sparsely observed regions.

3D Gaussian Splatting has been widely adopted for human avatar reconstruction due to its efficient optimization and real-time rendering. 
Earlier methods primarily targeted multi-view capture~\cite{li2024animatablegaussians}, monocular-video reconstruction~\cite{chen2025d3}, or mesh-guided settings, leveraging explicit Gaussians to model dynamic appearance, non-rigid deformation, and expressive whole-body motion.
More recently, single-image approaches~\cite{sun2026ani3dhuman} have substantially reduced the acquisition burden. LHM~\cite{qiu2025lhm} directly predicts a canonical animatable Gaussian avatar from one image in a feed-forward manner, while PERSONA~\cite{sim2025persona} leverages pose-diverse synthesized observations to optimize a personalized avatar. DynaAvatar~\cite{kwon2026dynaavatar} further incorporates motion-dependent garment deformation into zero-shot single-image reconstruction, demonstrating the potential of learned human and motion priors for generating animatable avatars from highly limited visual input. 

These methods demonstrate the potential of 3D Gaussian Splatting for high-fidelity, real-time human rendering. However, most existing methods still represent the body and clothing as a unified avatar, entangling garment geometry, skinning, appearance, and motion with the body. This limits garment-level control for transfer, virtual try-on, category-specific dynamics, and natural body--garment composition. Forwardrobe instead explicitly separates garments as independent Gaussian assets with dedicated geometry, appearance, and deformation properties.

\subsection{Garment-Aware Avatars}

Garment-aware avatars require clothing to be explicitly separated from the body and transferable across identities, shapes, and poses. Early mesh-based methods~\cite{bhatnagar2019multi,pons2017clothcap} introduced controllable body--garment representations and clothing retargeting, while MPMAvatar~\cite{lee2026mpmavatar} and PhysAvatar~\cite{zheng2024physavatar} modeled dynamic garments from multi-view observations. LayGA~\cite{lin2024layga} and Gaussian Wardrobe~\cite{chen2026gaussianwardrobe} further represented clothing as reusable layered Gaussian assets from multi-view videos.
LayerAvatar~\cite{zhang2025disentangled} learns a feed-forward diffusion model for generating component-disentangled Gaussian avatars and supports component transfer, while MonoCloth~\cite{jin2026monocloth} extends cloth-decoupled reconstruction to monocular video.
Disco4D~\cite{pang2025disco4d} separates and animates clothing Gaussians from a single image, but requires per-instance optimization rather than feed-forward inference.

Forwardrobe shares the goal of explicit garment modeling but addresses the more challenging single-image setting. Starting from a feed-forward Gaussian avatar, it identifies garment Gaussians in canonical space and represents clothing as an independent 3D asset with its own geometry, semantics, appearance, and skinning parameters. This requires stronger human priors and category-aware regularization to resolve the ambiguity caused by missing views and motion cues.

Unlike existing image-based 2D virtual try-on methods that primarily synthesize appearance in the image space, Forwardrobe reconstructs an editable, animatable, and re-renderable 3D garment asset, enabling novel-view rendering, cross-avatar garment transfer, and 3D virtual try-on.

\newcommand{\yes}{\textcolor{green!50!black}{\ding{51}}}
\newcommand{\no}{\textcolor{red!75!black}{\ding{55}}}

\begin{table}[!htbp]
\centering
\small
\renewcommand{\arraystretch}{1.15}
\setlength{\tabcolsep}{6.5pt}
\begin{tabular}{lcccc}
\toprule
\textbf{Method}
& \textbf{Single Img.}
& \textbf{FF.}
& \textbf{Dyn.}
& \textbf{VTON} \\
\midrule
LayGA                 & \no  & \no  & \yes & \yes \\
MonoCloth             & \no  & \no  & \yes & \yes \\
\midrule
LHM                   & \yes & \yes & \no  & \no  \\
PERSONA               & \yes & \no  & \yes & \no  \\
Disco4D               & \yes & \no  & \yes & \no  \\
DynaAvatar            & \yes & \yes & \yes & \no  \\
\midrule
\textbf{Forwardrobe (Ours)}
                      & \yes & \yes & \yes & \yes \\
\bottomrule
\end{tabular}

\caption{\textbf{Comparison with representative avatar reconstruction methods.}
``Single Img.'' denotes reconstruction from a single image;
``FF.'' indicates feed-forward inference without subject-specific optimization;
``Dyn.'' denotes motion-dependent non-rigid garment dynamics; and
``VTON'' denotes virtual try-on.}
\label{tab:method_comparison}
\end{table}

\begin{figure}[!htbp]
    \centering
    \includegraphics[width=0.95\linewidth]{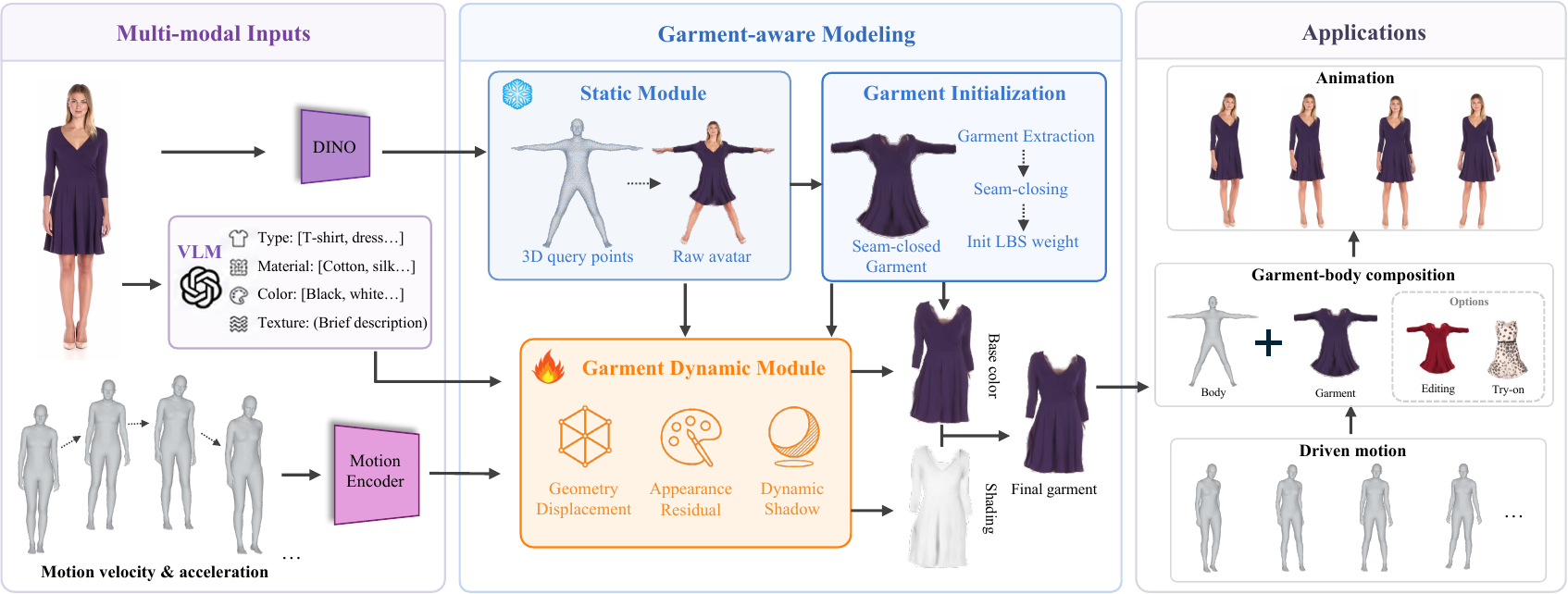}
    \caption{\textbf{Overview of Forwardrobe.}
    \textbf{1)} Given a single image and consecutive SMPL-X poses, we extract image features, VLM-derived garment descriptions, and temporal motion features.
    \textbf{2)} We construct an independent 3D garment asset through garment extraction, semantics-aware seam closing, skinning initialization, and base-appearance decomposition. The Garment Dynamic Module further predicts motion-dependent geometry, appearance, and shading residuals.
    \textbf{3)} The animated garment is composed with the body avatar to support animation, garment editing, and 3D virtual try-on.}
    \label{fig:pipeline}
\end{figure}

\section{Method}

\subsection{Overview}
\label{sec:overview}

Given a single image $I$ of a clothed person, Forwardrobe reconstructs an independent and animatable 3D Gaussian garment asset, which can subsequently be driven by arbitrary SMPL-X~\cite{pavlakos2019expressive} motion sequences.
As shown in Fig.~\ref{fig:pipeline}, a pretrained static reconstruction model provides the initial clothed avatar, while a visual language model (VLM) extracts structured garment descriptions. Forwardrobe then identifies the garment Gaussians and initializes their category-aware geometry, skinning weights, and base appearance. A Garment Dynamic Module predicts motion-dependent geometry, appearance, and shading variations. Finally, the animated garment is composed with the body Gaussians for animation, editing, and 3D virtual try-on.

\subsection{Garment-Aware Gaussian Avatar Representation}

\paragraph{Static initialization.}
We represent the reconstructed clothed avatar using a garment-aware Gaussian representation that explicitly separates the garment from the underlying body. The input static avatar consists of a set of 3D Gaussians anchored to a canonical-pose SMPL-X mesh. Let $\mathbf{q}_i$ denote the canonical sampling point on the SMPL-X surface, and $\Delta \mathbf{x}^{0}_i$ denotes the corresponding identity-dependent positional offset predicted by the static reconstruction model. The canonical center of the $i$-th Gaussian is therefore defined as $\mathbf{x}^{0}_i = \mathbf{q}_i + \Delta \mathbf{x}^{0}_i.$

Each Gaussian is parameterized as:
\begin{equation}
    \mathcal{G}^{0}_i =
    \left\{
    \mathbf{q}_i,
    \Delta \mathbf{x}^{0}_i,
    \mathbf{r}^{0}_i,
    \mathbf{s}^{0}_i,
    \alpha^{0}_i,
    \mathbf{c}^{0}_i,
    \mathbf{w}^{0}_i
    \right\},
\end{equation}
where $\mathbf{r}^{0}_i$, $\mathbf{s}^{0}_i$, $\alpha^{0}_i$, $\mathbf{c}^{0}_i$, and $\mathbf{w}^{0}_i$ denote its rotation, scale, opacity, color, and skinning weights, respectively. 

\paragraph{Garment decomposition.}
The Garment Labeling Module predicts per-Gaussian garment membership $m_i \in [0,1]$ from Gaussian and image features, supervised by observation-space and canonical-view garment masks.
In the observation space, we obtain a 2D garment mask $M^{g}_{\mathrm{img}}$ using an off-the-shelf human parsing and garment segmentation model~\cite{khirodkar2026sapiens2}. Each canonical Gaussian is transformed by the input pose $\boldsymbol{\theta}_0$ and projected onto the image plane:
\begin{equation}
    \mathbf{u}^{\mathrm{img}}_i
    =
    \Pi_{\mathrm{img}}
    \left(
        \mathrm{LBS}
        \left(
            \mathbf{x}^{0}_i,
            \mathbf{w}^{0}_i,
            \boldsymbol{\theta}_0
        \right)
    \right),
\end{equation}
The image-space cue is obtained by sampling the mask:
\begin{equation}
    m^{\mathrm{img}}_i
    =
    M^{g}_{\mathrm{img}}
    \left(
        \mathbf{u}^{\mathrm{img}}_i
    \right).
\end{equation}

Since the input-view mask can be unreliable in self-occluded regions, we additionally render the canonical avatar from front and back views and apply the same segmentation model to obtain
$\{M^{g}_{\mathrm{front}}, M^{g}_{\mathrm{back}}\}$. The corresponding canonical-space cues are computed analogously by projecting each Gaussian onto these views and sampling the masks.

Supervised by these multi-view masks, a classifier
$f_{\mathrm{label}}$ predicts the garment membership $m_i$.
We then partition the initial avatar into garment and body layers:
\begin{equation}
    \mathcal{I}^{g}
    =
    \left\{
        i \mid m_i > \eta_g
    \right\},
    \qquad
    \mathcal{G}^{g}
    =
    \left\{
        \mathcal{G}^{0}_i
        \mid i \in \mathcal{I}^{g}
    \right\},
\end{equation}
where $\eta_g=0.5$ is the garment membership threshold, and the remaining Gaussians constitute the body layer $\mathcal{G}^{b}$.

\paragraph{Gaussian garment asset.}
We organize the decomposed garment Gaussians into an independent 3D garment asset. Each garment Gaussian is parameterized as:
\begin{equation}
    \mathcal{G}^{g}_i =
    \left\{
        \mathbf{q}_i,
        \Delta \mathbf{x}^{g}_i,
        \mathbf{r}^{g}_i,
        \mathbf{s}^{g}_i,
        \alpha^{g}_i,
        \mathbf{c}^{\mathrm{SH},g}_i,
        \mathbf{w}^{g}_i
    \right\},
    \qquad i \in \mathcal{I}^{g},
\end{equation}
where $\Delta \mathbf{x}^{g}_i$ denotes the garment-specific offset from the sampled SMPL-X point $\mathbf{q}_i$, $\mathbf{c}^{\mathrm{SH},g}_i$ denotes the static spherical-harmonic appearance, and $\mathbf{w}^{g}_i$ denotes the garment-specific skinning weights. The canonical center of the garment Gaussian is defined as: $\mathbf{x}^{g}_i
    =
    \mathbf{q}_i + \Delta \mathbf{x}^{g}_i.$

The skinning weights provide coarse pose-driven deformation, while the static SH coefficients encode an albedo-like base appearance of the garment. Since motion-dependent appearance changes are dominated by illumination rather than intrinsic color variation, the Garment Dynamic Module models them using bounded SH residuals and a pose-dependent darkening gain.

A VLM additionally extracts a structured garment description $\boldsymbol{\tau}$, including category, texture, color, and material. These language-level cues guide category-aware garment continuity initialization and dynamic modeling. The complete garment asset is represented as:
\begin{equation}
    \mathcal{A}^{g}
    =
    \left(
        \left\{
            \mathcal{G}^{g}_i
        \right\}_{i \in \mathcal{I}^{g}},
        \boldsymbol{\tau}
    \right).
\end{equation}

\paragraph{Garment continuity initialization.}

Body-aligned clothing can inherit the original template support, but a skirt should span the lower-body midline instead of following two disconnected leg regions. According to the VLM-derived category $\boldsymbol{\tau}$, we redistribute inner-thigh garment Gaussians toward the canonical mid-plane:
\begin{equation}
    \tilde{\mathbf{x}}_i
    =
    \mathbf{x}^{g}_i
    +
    \gamma_i
    \left(
        \Pi_{\mathrm{mid}}(\mathbf{x}^{g}_i)
        -
        \mathbf{x}^{g}_i
    \right),
\end{equation}
where $\Pi_{\mathrm{mid}}(\cdot)$ projects onto the body mid-plane and $\gamma_i \in [0,1]$ decreases with the distance to the seam. This converts the leg-separated initialization into a continuous canonical garment representation.
We define the base center used by every subsequent module as:
\begin{equation}
 \bar{\mathbf{x}}^{g}_i=
 \begin{cases}
 \tilde{\mathbf{x}}_i, & \text{if } \boldsymbol{\tau}\text{ is skirt-like},\\
 \mathbf{x}^{g}_i, & \text{otherwise}.
 \end{cases}
 \label{eq:base_center}
\end{equation}

Seam closing alone is insufficient because the primitives can retain incompatible left/right-leg weights. We therefore query a continuous diffused skinning field $\mathcal{W}:\mathbb{R}^{3}\rightarrow\Delta^{J-1}$~\cite{kwon2026dynaavatar} at $K=13$ samples over each anisotropic Gaussian:
\begin{equation}
    \hat{\mathbf{w}}_i^g
    =
    \operatorname{Normalize}
    \left(
        \sum_{k=1}^{K}
        \alpha_k
        \mathcal{W}
        \left(
            \tilde{\mathbf{x}}_i
            +
            \mathbf{R}_i
            \operatorname{diag}(\mathbf{s}_i)
            \boldsymbol{\delta}_k
        \right)
    \right),
\end{equation}
where $\hat{\mathbf{w}}_i^g$ denotes the initialized garment skinning weights, computed by Eq.~(9) for skirt-like garments and set to the original garment weights $\mathbf{w}_i^g$ otherwise.
Here, $\boldsymbol{\delta}_k$ is the $k$-th local sample offset, and $\alpha_k$ is its aggregation weight. The Gaussian rotation $\mathbf{R}_i$ and scale $\mathbf{s}_i$ transform the samples from the normalized local frame to canonical space.

\begin{figure}[!htbp]
    \centering
    \includegraphics[width=0.82\linewidth]{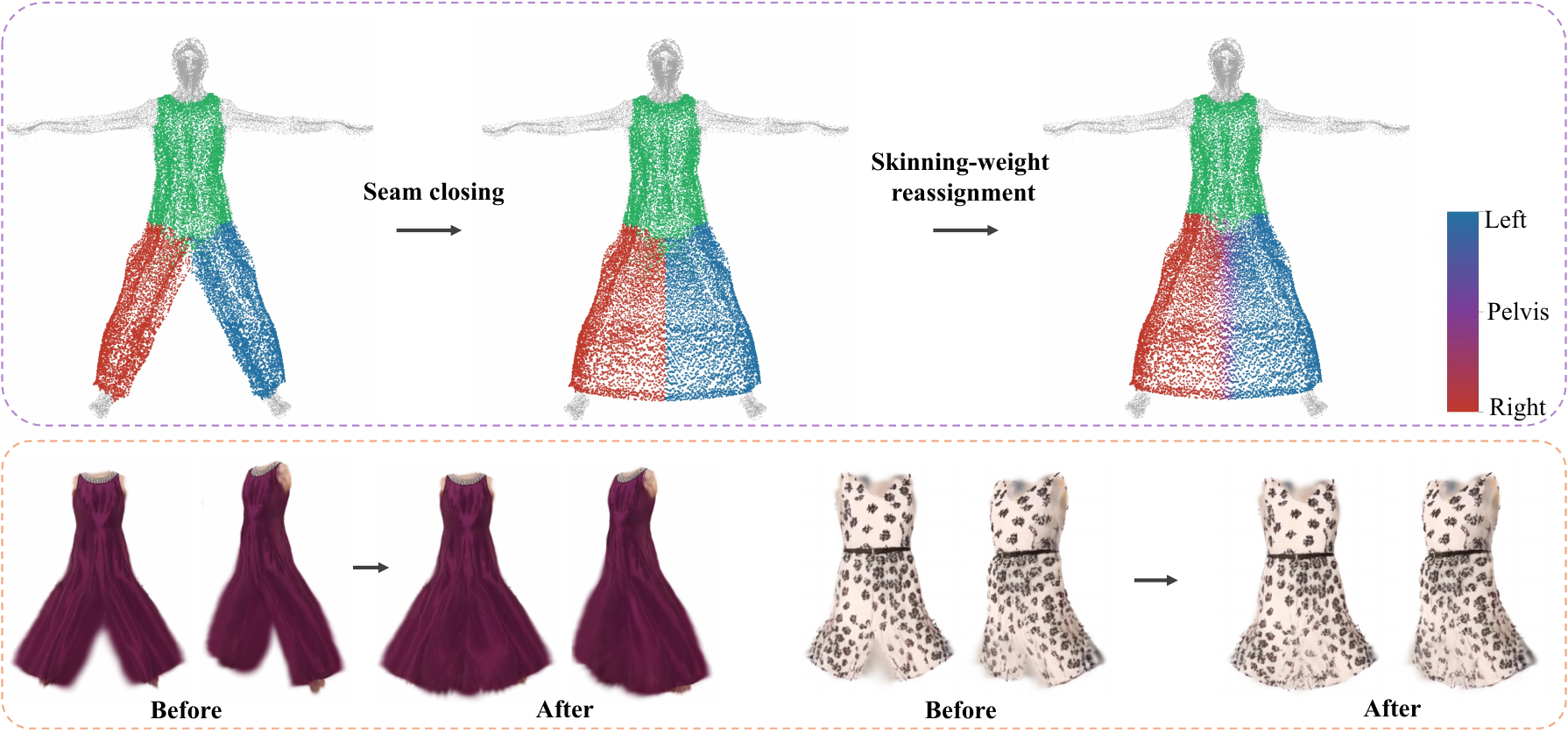}
    \caption{\textbf{Garment continuity initialization.}
\textbf{Top:} seam closing followed by diffused skinning-weight reassignment.
\textbf{Bottom:} improved spatial continuity under animation.}
    \label{fig:seam_close}
\end{figure}

\subsection{Garment Dynamic Module}
\label{sec:garment_dynamic}

The static garment asset cannot fully capture pose-dependent non-rigid deformation and appearance variation, particularly for loose garments. We therefore introduce a residual-based Garment Dynamic Module that predicts motion-conditioned updates to Gaussian geometry and appearance while preserving the continuity-aware initialization.

\paragraph{Dynamic conditioning.}
For each garment Gaussian, we construct a local asset feature $\mathbf{g}_i$ from its base position $\bar{\mathbf{x}}^{g}_i$, Gaussian attributes, and learned local feature $\mathbf{z}^{g}_i$. 
Garment labels restrict garment-specific prediction and supervision to the selected Gaussian subset.

Given a target frame $t$, a temporal motion encoder processes a short SMPL-X motion window $\mathcal{M}_t$ centered at the target pose:
$\mathbf{f}^{\mathrm{mot}}_t
    =
    E_{\mathrm{mot}}(\mathcal{M}_t).$
The encoded feature captures articulated pose together with temporal motion cues, such as joint velocity and acceleration. We additionally extract source-image features:
$\mathbf{f}^{\mathrm{img}}=E_{\mathrm{DINO}}(I)$
to preserve subject- and garment-specific visual information~\cite{oquab2023dinov2}. Our architecture also incorporates a VLM-derived material feature $\mathbf{f}^{\mathrm{sem}}$, which describes garment-level semantics such as category, texture, and material.

These multimodal conditions are fused with the per-Gaussian asset feature by a lightweight transformer--MLP hybrid adapter:
\begin{equation}
    \mathbf{h}_{i,t}
    =
    A_{\mathrm{dyn}}
    \left(
        \mathbf{g}_i,\,
        \mathbf{f}^{\mathrm{mot}}_t,\,
        \mathbf{f}^{\mathrm{img}},\,
        \mathbf{f}^{\mathrm{sem}}
    \right).
\end{equation}

\paragraph{Residual garment adaptation.}
Given the fused feature $\mathbf{h}_{i,t}$, the dynamic head predicts residuals for garment geometry and appearance:
\begin{equation}
    \left(
        \Delta\mathbf{x}_{i,t},
        \Delta\alpha_{i,t},
        \Delta\mathbf{s}_{i,t},
        \Delta\mathbf{r}_{i,t},
        \Delta\mathbf{c}^{\mathrm{SH}}_{i,t},
        \rho_{i,t}
    \right)
    =
    H_{\mathrm{dyn}}(\mathbf{h}_{i,t}),
\end{equation}
which refine position, opacity, scale, rotation, SH appearance, and pose-dependent shading, respectively. Residual prediction preserves the static garment asset while capturing motion-dependent variations.

We further adapt the initialized skinning weights using a lightweight, time-independent adapter:
\begin{equation}
\begin{aligned}
 \Delta\boldsymbol{\ell}^{w}_i
 &=F_{\rm skin}(\bar{\mathbf{x}}^{g}_i,\log\mathbf{s}_i,
 \hat{\mathbf{w}}^{g}_i,m_i),\\
 \mathbf{w}_i
 &=\operatorname{Softmax}\!\left[
 \log(\hat{\mathbf{w}}^{g}_i+\epsilon)
 +\lambda_w\Delta\boldsymbol{\ell}^{w}_i
 \right].
\end{aligned}
\end{equation}

The position residual is applied in canonical space by:
\begin{equation}
 \mathbf{x}^{c}_{i,t}=\bar{\mathbf{x}}^{g}_i+
 \lambda_x\Delta\mathbf{x}_{i,t},\qquad
 \mathbf{x}_{i,t}=\operatorname{LBS}
 (\mathbf{x}^{c}_{i,t},\mathbf{w}_i,\boldsymbol{\theta}_t).
 \label{eq:dynamic_center}
\end{equation}

The remaining residuals refine the corresponding static Gaussian attributes, preserving the continuity-aware canonical initialization while introducing motion-dependent geometric and appearance details.

\paragraph{Dynamic garment shading.}
Garment appearance variations during animation can arise from pose-dependent shading caused by folds, self-occlusion, and cast shadows. We therefore retain the static SH coefficients as a stable base appearance and model temporal changes with bounded residuals rather than unconstrained time-varying colors.
A small SH residual accounts for remaining appearance variations:
\begin{equation}
    \widetilde{\mathbf{c}}^{\mathrm{SH}}_{i,t}
    =
    \mathbf{c}^{\mathrm{SH}}_{i}
    +
    \lambda_{\mathrm{SH}}
    \Delta\mathbf{c}^{\mathrm{SH}}_{i,t}.
\end{equation}

To model the dominant shading variation, the dynamic shadow head predicts a raw response $\rho_{i,t}$, which is converted into a non-negative shadow strength $d_{i,t}$ and a bounded darkening gain $g_{i,t}$:
\begin{equation}
\begin{aligned}
    d_{i,t}
    &=
    \operatorname{softplus}
    \left(
        \rho_{i,t}-b_{\mathrm{sh}}
    \right),\\
    g_{i,t}
    &=
    \operatorname{clamp}
    \left(
        e^{-\lambda_{\mathrm{sh}}d_{i,t}},
        g_{\min},
        1
    \right).
\end{aligned}
\end{equation}
The final SH coefficients used for rendering are
\begin{equation}
    \mathbf{c}^{\mathrm{SH,final}}_{i,t}
    =
    g_{i,t}
    \widetilde{\mathbf{c}}^{\mathrm{SH}}_{i,t}.
\end{equation}

Since $g_{i,t}\in[g_{\min},1]$, this branch models bounded pose-dependent darkening while preventing arbitrary color shifts and excessive brightening.

\paragraph{Garment--body composition.}
Since dynamic garment offsets are applied only to garment Gaussians, the garment may separate from the LBS-driven body near shared boundaries such as the collar, cuffs, and waist. We therefore introduce a lightweight, training-free composition step that identifies nearby garment--body junctions in canonical space and smoothly propagates garment displacements to adjacent body Gaussians before LBS. 
These operations
preserve coherent boundaries during animation and improve local
garment--body integration in virtual try-on without additional learnable
parameters.

\begin{figure}[!htbp]
    \centering
    \includegraphics[width=0.95\linewidth]{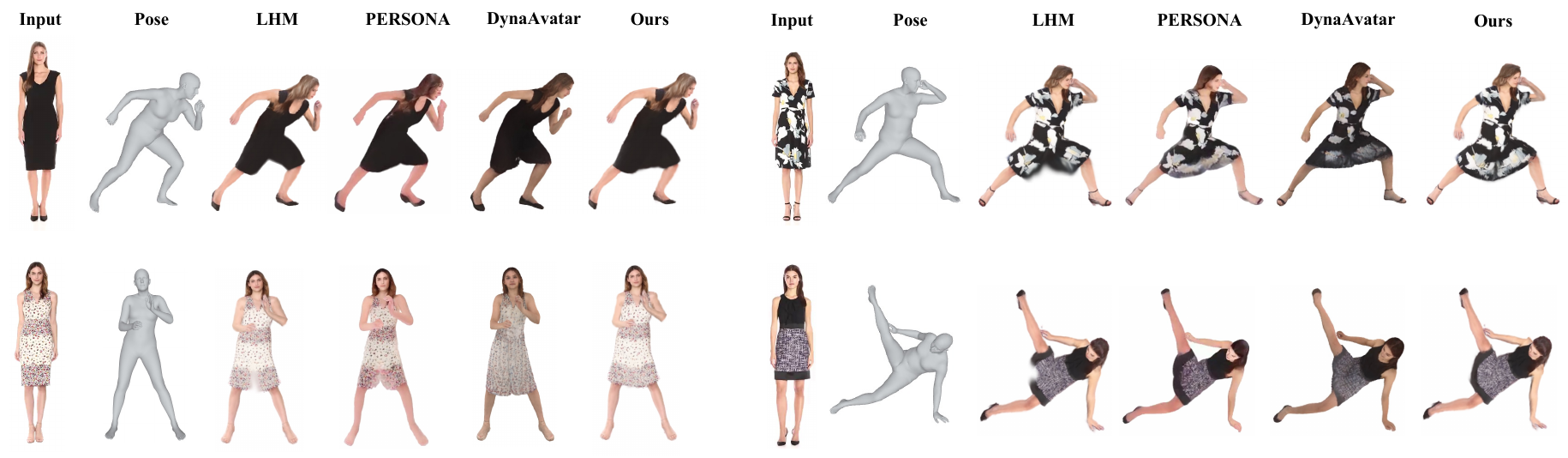}
    \caption{\textbf{Qualitative comparison under challenging target poses.}
Given a single input image and a target SMPL-X pose, we compare LHM, PERSONA, DynaAvatar, and Forwardrobe. Our method better preserves garment appearance and structural continuity while producing more coherent deformations, particularly for loose garments under large articulated motions.}
    \label{fig:qualitative}
\end{figure}

\begin{table}[!htbp]
    \centering
    \setlength{\tabcolsep}{3.2pt}
    \renewcommand{\arraystretch}{1.05}
    \resizebox{0.94\textwidth}{!}{
    \begin{tabular}{lcccccccccccc}
        \toprule
        \multirow{3}{*}{Method}
        & \multicolumn{6}{c}{UBC Fashion}
        & \multicolumn{6}{c}{NeuMan} \\
        \cmidrule(lr){2-7}
        \cmidrule(lr){8-13}

        & \multicolumn{3}{c}{Full Image}
        & \multicolumn{3}{c}{Garment Region}
        & \multicolumn{3}{c}{Full Image}
        & \multicolumn{3}{c}{Garment Region} \\
        \cmidrule(lr){2-4}
        \cmidrule(lr){5-7}
        \cmidrule(lr){8-10}
        \cmidrule(lr){11-13}

        & PSNR $\uparrow$
        & SSIM $\uparrow$
        & LPIPS $\downarrow$
        & PSNR $\uparrow$
        & SSIM $\uparrow$
        & LPIPS $\downarrow$
        & PSNR $\uparrow$
        & SSIM $\uparrow$
        & LPIPS $\downarrow$
        & PSNR $\uparrow$
        & SSIM $\uparrow$
        & LPIPS $\downarrow$ \\
        \midrule

        IDOL
        & 20.31 & 0.848 & 0.111 & 13.42 & 0.379 & 0.061
        & 20.94 & 0.923 & 0.074 & 14.44 & 0.630 & 0.049 \\

        PERSONA
        & 18.36 & 0.806 & 0.133 & 15.83 & 0.479 & 0.058
        & 18.03 & 0.914 & 0.117 & 13.29 & 0.597 & 0.051 \\

        LHM 
        & \underline{23.02} & \underline{0.877} & \textbf{0.064} & \underline{17.54} & 0.574 & \underline{0.032} 
        & 20.06 & 0.924 & 0.069 & 13.73 & 0.660 & 0.043 \\

        DynaAvatar
        & 20.51 & 0.832 & 0.079 & 14.23 & 0.427 & 0.041
        & 20.39 & \underline{0.927} & \underline{0.063} & 12.79 & 0.639 & 0.043 \\

        \midrule

        Ours (500M)
        & 22.85 & 0.875 & 0.065 & 17.44 & \underline{0.586} & \underline{0.032}
        & \underline{21.06} & 0.926 & 0.066 & \underline{15.12} & \textbf{0.738} & \underline{0.041} \\

        Ours (1B)
        & \textbf{23.19} & \textbf{0.885} & \textbf{0.064} & \textbf{18.30} & \textbf{0.604} &  \textbf{0.031}
        & \textbf{21.58} & \textbf{0.930} & \textbf{0.062} & \textbf{15.30} & \underline{0.730} & \textbf{0.040} \\

        \bottomrule
    \end{tabular}
    }
    \caption{\textbf{Quantitative comparison on UBC Fashion and NeuMan.}
    We report PSNR, SSIM, and LPIPS over the full image and the masked garment region. The best and second-best results are highlighted in boldface and underlined, respectively.}
    \label{tab:quantitative}
\end{table}

\subsection{Training Losses}
\label{sec:losses}

We train Forwardrobe with reconstruction, garment deformation
regularization, and appearance supervision:
\begin{equation}
    \mathcal{L}
    =
    \mathcal{L}_{\mathrm{rec}}
    +
    \mathcal{L}_{\mathrm{reg}}
    +
    \mathcal{L}_{\mathrm{app}}.
\end{equation}

\paragraph{Garment-focused reconstruction.}
The reconstruction objective $\mathcal{L}_{\mathrm{rec}}$ combines
full-image and foreground photometric losses with garment-masked RGB,
image-gradient, and perceptual supervision. Global and garment-focused
LPIPS losses preserve overall visual fidelity and perceptual garment
details, while silhouette supervision maintains accurate foreground
boundaries. We additionally use weak image-space anchors to prevent
dynamic refinement from unnecessarily degrading reliable regions of
the initialized avatar.

\paragraph{Garment deformation regularization.}
To encourage coherent and stable garment deformation, we construct a
$K$-nearest-neighbor ($K=8$) graph over the canonical Gaussian centers and
retain garment--garment edges to regularize the predicted dynamic
offsets:
\begin{equation}
\begin{aligned}
    \mathcal{L}_{\mathrm{reg}}
    ={}&
    \lambda_{\mathrm{lap}}\mathcal{L}_{\mathrm{lap}}
    +
    \lambda_{\mathrm{edge}}\mathcal{L}_{\mathrm{edge}}
    +
    \lambda_{\mathrm{temp}}\mathcal{L}_{\mathrm{temp}}
    +
    \lambda_{\mathrm{spr}}\mathcal{L}_{\mathrm{spr}}
    \\
    &+
    \lambda_{\mathrm{geom}}\mathcal{L}_{\mathrm{geom}}
    +
    \lambda_{\mathrm{anchor}}\mathcal{L}_{\mathrm{anchor}} .
\end{aligned}
\end{equation}
The $\mathcal{L}_{\mathrm{lap}}$ and $\mathcal{L}_{\mathrm{edge}}$ encourage locally coherent Gaussian
motion, while the $\mathcal{L}_{\mathrm{temp}}$ penalizes abrupt acceleration and
reduces animation jitter. The spring term penalizes inconsistent
offsets between neighboring Gaussians. We further preserve the
seam-closed garment geometry using an asymmetric outer-boundary
constraint together with bounded position and skinning corrections.

\paragraph{Residual shading supervision.}
Let $\mathbf{I}^{\mathrm{base}}_t$ denote the base-appearance rendering before applying dynamic shading. We derive a darken-only pseudo target from the target-to-base luminance ratio and supervise the predicted shading only in reliable garment regions:
\begin{equation}
\begin{aligned}
    \mathbf{S}^{*}_t
    &=
    \operatorname{clamp}
    \left(
        \frac{Y(\mathbf{I}_t)}
        {\max(Y(\mathbf{I}^{\mathrm{base}}_t),\epsilon)},
        s_{\min},1
    \right),\\
    \mathcal{L}_{\mathrm{shade}}
    &=
    \frac{
        \sum_{\mathbf{u}}
        \mathbf{R}_t(\mathbf{u})
        \operatorname{SmoothL1}
        \left(
            \hat{\mathbf{S}}_t(\mathbf{u}),
                \mathbf{S}^{*}_t(\mathbf{u})
        \right)
    }{
        \sum_{\mathbf{u}}
        \mathbf{R}_t(\mathbf{u})+\epsilon
    } .
\end{aligned}
\end{equation}

Here, $Y(\cdot)$ denotes luminance, $\hat{\mathbf{S}}_t$ is the predicted shading gain, and $\mathbf{u}$ indexes image pixels. The reliability mask $\mathbf{R}_t$ excludes uncertain boundaries, dark regions, and high-gradient textures, preventing garment patterns and base-appearance errors from entering the shading branch. 
The overall appearance objective $\mathcal{L}_{\mathrm{app}}$ further includes temporal shading regularization and constraints on the magnitude and spatial smoothness of the SH correction.

\section{Experiments}

\subsection{Experimental Setup}
\label{sec:experimental_setup}

\paragraph{Datasets and protocol.}
We train Forwardrobe on UBC Fashion~\cite{zablotskaia2019dwnet}, X-Humans~\cite{shen2023xhumans}, and NeuMan~\cite{jiang2022neuman}, which provide complementary training data. UBC Fashion contains diverse loose garments and fashion-oriented motions, X-Humans contributes detailed body-aligned clothing, and NeuMan introduces in-the-wild scenes with moving cameras and complex backgrounds. We follow the official test splits for evaluation. On UBC Fashion, we evaluate 50 test subjects and uniformly sample 10 target frames from each sequence, using one reference frame as input. All evaluation identities and frames are held out from training.

\paragraph{Metrics.}
We evaluate Forwardrobe against representative single-image avatar methods, including LHM~\cite{qiu2025lhm}, IDOL~\cite{zhuang2025idol}, PERSONA~\cite{sim2025persona}, and DynaAvatar~\cite{kwon2026dynaavatar}. 
For fair comparison, all predictions and reference frames are transformed to the same body-centered canvas using the fitted camera and SMPL-X parameters, resized to a fixed scale, and white-padded, thereby excluding the large blank regions in the original NeuMan frames.
We report PSNR, SSIM, and LPIPS~\cite{zhang2018lpips} on the normalized canvas, together with the same metrics computed within the garment region to better assess garment reconstruction quality.

\paragraph{Implementation details.}

We use LHM-1B as the static backbone and Qwen2-VL-7B~\cite{wang2024qwen2} to extract garment descriptions. Each avatar contains 40K Gaussian primitives. Forwardrobe is trained for 50K iterations on two NVIDIA RTX 4090 GPUs. At inference time, it reconstructs a garment-aware avatar in approximately 10 seconds on a single GPU and renders each frame in 65–145 ms.

\subsection{Comparison with State-of-the-Art Methods}

Tab.~\ref{tab:quantitative} and Fig.~\ref{fig:qualitative} present quantitative and qualitative comparisons between Forwardrobe and state-of-the-art methods. 
The reported LHM results are obtained using its strongest LHM-1B variant.
Since DynaAvatar is only available with the LHM-500M backbone, we use the same static module for Forwardrobe in this comparison.
Results on UBC Fashion and NeuMan show that Forwardrobe achieves competitive overall rendering quality, with particularly strong performance in the garment regions emphasized by our design. 

We additionally evaluate the reconstructed avatars using challenging motion sequences in the qualitative comparison.
As shown in Fig.~\ref{fig:qualitative}, LHM closely inherits the leg-separated SMPL-X topology and often treats skirts as loosely fitted pants, leading to visible hem splitting under large leg motions. PERSONA relies on 2D generative priors to synthesize pose-diverse supervision, which can introduce inconsistencies in garment topology, facial identity, and clothing patterns. 
For example, the first-row subject shows noticeable appearance distortion and blurred facial details. 
DynaAvatar explicitly models dynamic clothing, but its learned motion and garment priors may occasionally misinterpret garment structure, leading to artifacts such as spurious slits and loss of fine texture details, as observed in the two examples on the left.
In contrast, Forwardrobe leverages VLM-derived garment descriptions for category-conditioned garment continuity initialization, preserving garment continuity under challenging poses. Its modular design also allows stronger static reconstruction backbones to be integrated with minimal modification and overhead, further improving identity preservation and appearance fidelity.

\begin{figure}[!htbp]
    \centering
    \includegraphics[width=0.62\linewidth]{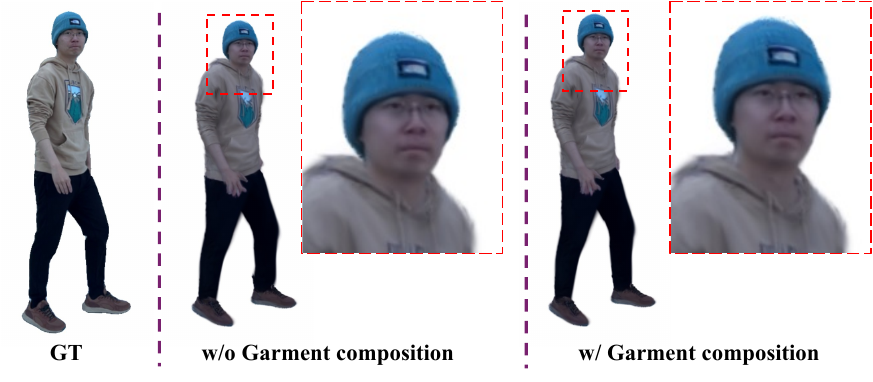}
    \caption{\textbf{Garment composition.}
    Our composition strategy preserves the static body appearance while integrating the dynamically animated garment.}
    \label{fig:ablation}
\end{figure}

\begin{table}[!htbp]
  \centering
  \small
  \setlength{\tabcolsep}{5pt}
  \begin{tabular}{lccc}
    \toprule
    Method                  & PSNR $\uparrow$       & SSIM $\uparrow$        & LPIPS $\downarrow$     \\
    \midrule
    w/o VLM-guided Init. & 20.96 & 0.921 & 0.072
\\
    w/o Dynamic & 20.83 & 0.921 & 0.069
 \\
    w/o Composition      & 21.48 & \underline{0.924} & 0.067
 \\
    w/o Shading    & \underline{21.55} & \underline{0.924} & \underline{0.066}
 \\

    Ours  &    
\textbf{21.78} & \textbf{0.925} & \textbf{0.065} \\
    \bottomrule
  \end{tabular}
  \caption{\textbf{Ablation study.}
We evaluate the contributions of continuity initialization, the
Garment Dynamic Module, garment--body composition, and shading decomposition.}
  \label{tab:ablation}
\end{table}

\subsection{Ablation Studies}

We evaluate the key components of Forwardrobe on four NeuMan subjects and four UBC Fashion subjects, as reported in Tab.~\ref{tab:ablation}.
Removing the VLM-guided continuity
initialization, including seam closing and skinning-weight
reassignment, substantially degrades reconstruction quality,
demonstrating its importance for loose garments. Specifically,
the VLM-derived garment classification determines whether
category-specific continuity initialization is activated; without
it, skirt-like garments fall back to the generic body-aligned
initialization.
Removing the Garment Dynamic Module causes the largest PSNR drop, confirming that LBS alone cannot adequately capture pose-dependent folds, draping, and other non-rigid garment deformations.

Garment composition further improves reconstruction quality by aligning
the independently driven body and garment layers. As shown in
Fig.~\ref{fig:ablation}, without composition, spatial misalignment may
produce unnatural transitions at body--garment interfaces, whereas our
strategy yields more coherent connections. Finally, shading
decomposition preserves reconstruction fidelity while separating stable
garment appearance from motion-dependent shading, enabling controllable
recoloring and appearance editing.

\subsection{Applications}

\begin{figure}[!htbp]
    \centering
    \includegraphics[width=0.68\linewidth]{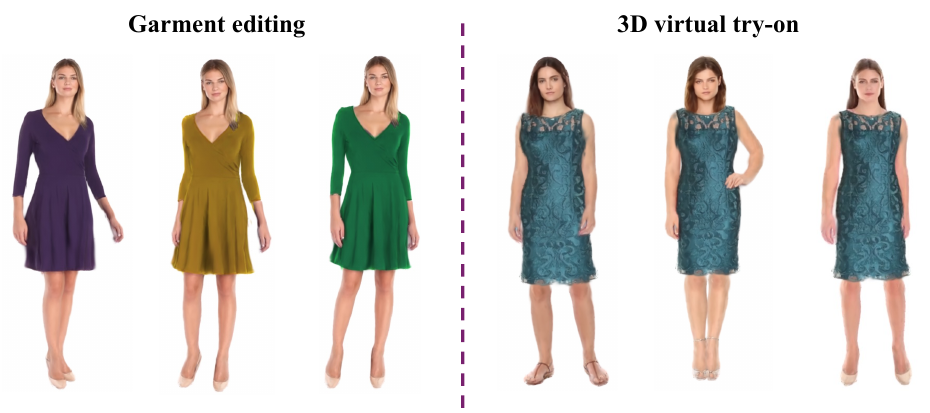}
    \caption{\textbf{Applications.}
    Forwardrobe supports garment editing and 3D virtual try-on. The resulting avatars preserve the edited or transferred garments and can be freely animated with arbitrary motion sequences.}

    \label{fig:application}
\end{figure}

Since Forwardrobe represents clothing as an independent 3D asset, it naturally supports garment-specific editing of the reconstructed avatar. One straightforward application is appearance editing, particularly garment recoloring. As shown in Fig.~\ref{fig:teaser} and Fig.~\ref{fig:application}, our approximate decomposition of garment appearance into a stable base component and shading enables the base color to be modified independently while preserving and reapplying the original shading effects, yielding more natural and visually coherent results. The edited garment remains fully animatable and can be rendered under arbitrary poses and viewpoints without sacrificing its geometric or appearance consistency.

Moreover, garment assets extracted from different subjects can be organized into a digital ``wardrobe.'' For virtual try-on, the target avatar's original garment primitives are removed, and a selected source garment is inserted through their shared canonical correspondence. The transferred asset retains its geometry, appearance, skinning, and motion-conditioned residual model, while the target avatar preserves its body appearance and driving motion. This enables 3D virtual try-on and garment replacement within a shared canonical template, without garment-specific retraining.

\section{Conclusion}

We presented Forwardrobe, a feed-forward framework that improves the reconstruction and animation of loose garments from a single clothed-person image. 
Specifically, category-aware geometry and skinning initialization improve the visual continuity of skirt-like garments, while bounded geometry and appearance residuals preserve garment quality under motion.
Unlike monolithic avatar representations, Forwardrobe separates the garment's geometry, appearance, skinning, and motion-conditioned residual model from the remaining body, enabling independent animation and recomposition.
The resulting assets support appearance editing, cross-avatar transfer, and 3D virtual try-on within a shared canonical template, moving single-image avatar reconstruction toward reusable digital wardrobes.

\bibliographystyle{unsrt}
\bibliography{references}

\end{document}